\documentclass[10pt,twocolumn,letterpaper]{article}

\usepackage[pagenumbers]{cvpr} 

\usepackage{graphicx}
\usepackage{amsmath}
\usepackage{amssymb}
\usepackage{booktabs}

\usepackage{multirow}
\usepackage{pifont}
\usepackage{threeparttable}

\usepackage[pagebackref,breaklinks,colorlinks]{hyperref}

\usepackage[capitalize]{cleveref}
\crefname{section}{Sec.}{Secs.}
\Crefname{section}{Section}{Sections}
\Crefname{table}{Table}{Tables}
\crefname{table}{Tab.}{Tabs.}

\usepackage[accsupp]{axessibility}  

\def\cvprPaperID{6382} 
\def\confName{CVPR}
\def\confYear{2023}

\begin{document}
\title{Adaptive Sparse Convolutional Networks with Global Context Enhancement for Faster Object Detection on Drone Images}

\author{
    \textbf{Bowei Du}$^{1,2\dag}$, \textbf{Yecheng Huang}$^{1,2\dag}$, \textbf{Jiaxin Chen}$^{2}$, \textbf{Di Huang}$^{1,2,3*}$\\
    {$^1$ State Key Laboratory of Software Development Environment, Beihang University, Beijing, China}\\
    {$^2$ School of Computer Science and Engineering, Beihang University, Beijing, China}\\
    {$^3$ Hangzhou Innovation Institute, Beihang University, Hangzhou, China}\\
    {\{boweidu, ychuang, jiaxinchen, dhuang\}@buaa.edu.cn}
}

\maketitle
\begin{abstract}
Object detection on drone images with low-latency is an important but challenging task on the resource-constrained unmanned aerial vehicle (UAV) platform. This paper investigates optimizing the detection head based on the sparse convolution, which proves effective in balancing the accuracy and efficiency. Nevertheless, it suffers from inadequate integration of contextual information of tiny objects as well as clumsy control of the mask ratio in the presence of foreground with varying scales. To address the issues above, we propose a novel global context-enhanced adaptive sparse convolutional network (CEASC). It first develops a context-enhanced group normalization (CE-GN) layer, by replacing the statistics based on sparsely sampled features with the global contextual ones, and then designs an adaptive multi-layer masking strategy to generate optimal mask ratios at distinct scales for compact foreground coverage, promoting both the accuracy and efficiency. Extensive experimental results on two major benchmarks, i.e. VisDrone and UAVDT, demonstrate that CEASC remarkably reduces the GFLOPs and accelerates the inference procedure when plugging into the typical state-of-the-art detection frameworks (e.g. RetinaNet and GFL V1) with competitive performance. Code is available at https://github.com/Cuogeihong/CEASC.
\end{abstract}

\renewcommand{\thefootnote}{}
\footnotetext{$\dag$ indicates equal contribution.}
\footnotetext{$*$ refers to the corresponding author.}

\section{Introduction}
\label{sec:intro}

Recent progress of deep neural networks (\emph{e.g.} CNNs and Transformers) has significantly boosted the performance of object detection on public benchmarks such as COCO~\cite{COCO2014}. By contrast, building detectors for unmanned aerial vehicle (UAV) platforms currently remains a challenging task. On the one hand, existing studies are keen on designing complicated models to reach high accuracies of tiny objects on high-resolution drone imagery, which are computationally consuming. On the other hand, the hardware equipped with UAVs is often resource-constrained, raising an urgent demand in lightweight deployed models for fast inference and low latency. 
\begin{figure}[!t]
    \centering
    \begin{subfigure}{\linewidth}
        \centering
        \includegraphics[width=1\linewidth]{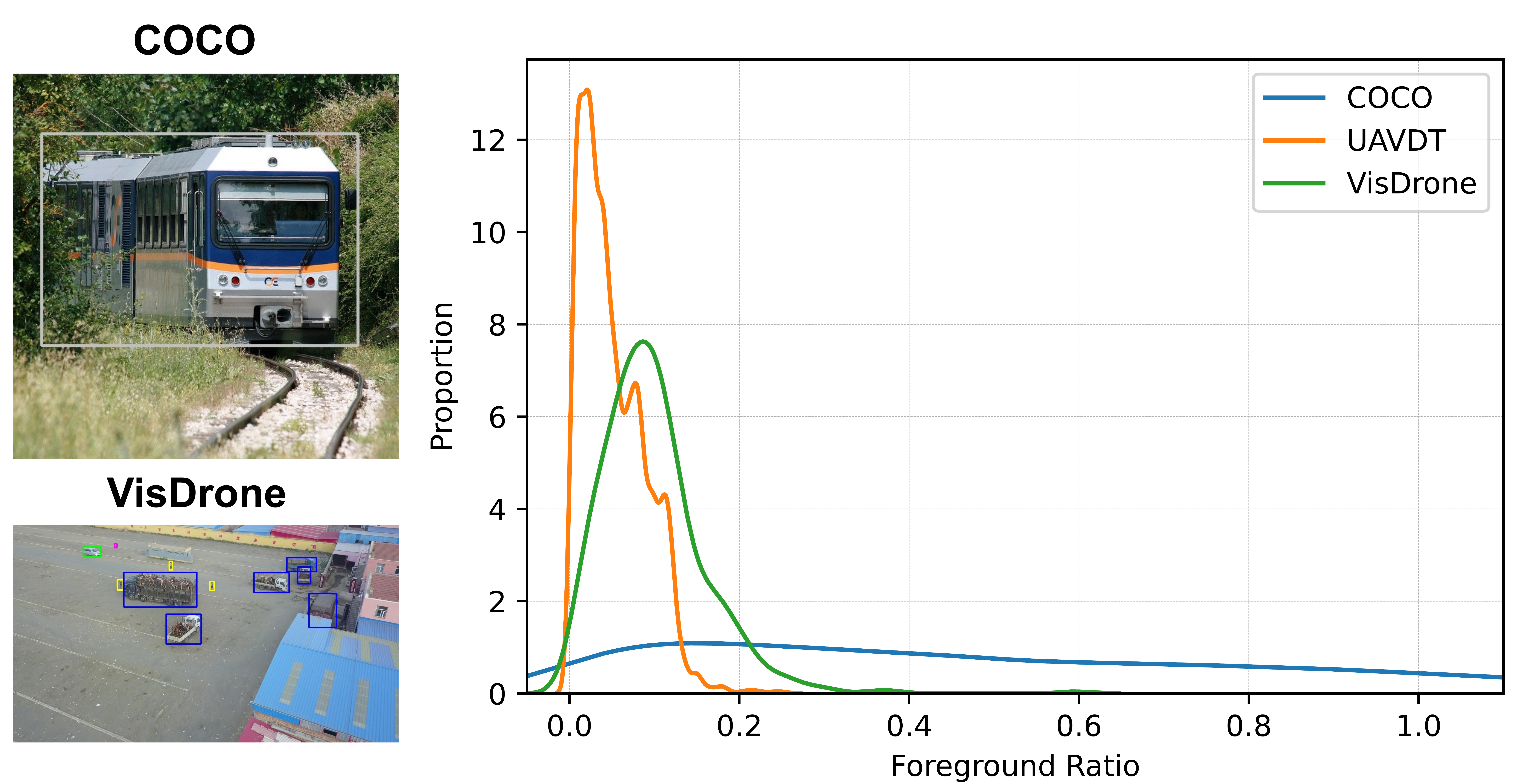}
        \caption{}
        \label{fig:short-a}
    \end{subfigure}
    \hfill
    \begin{subfigure}{\linewidth}
        \centering
    \includegraphics[width=1\linewidth]{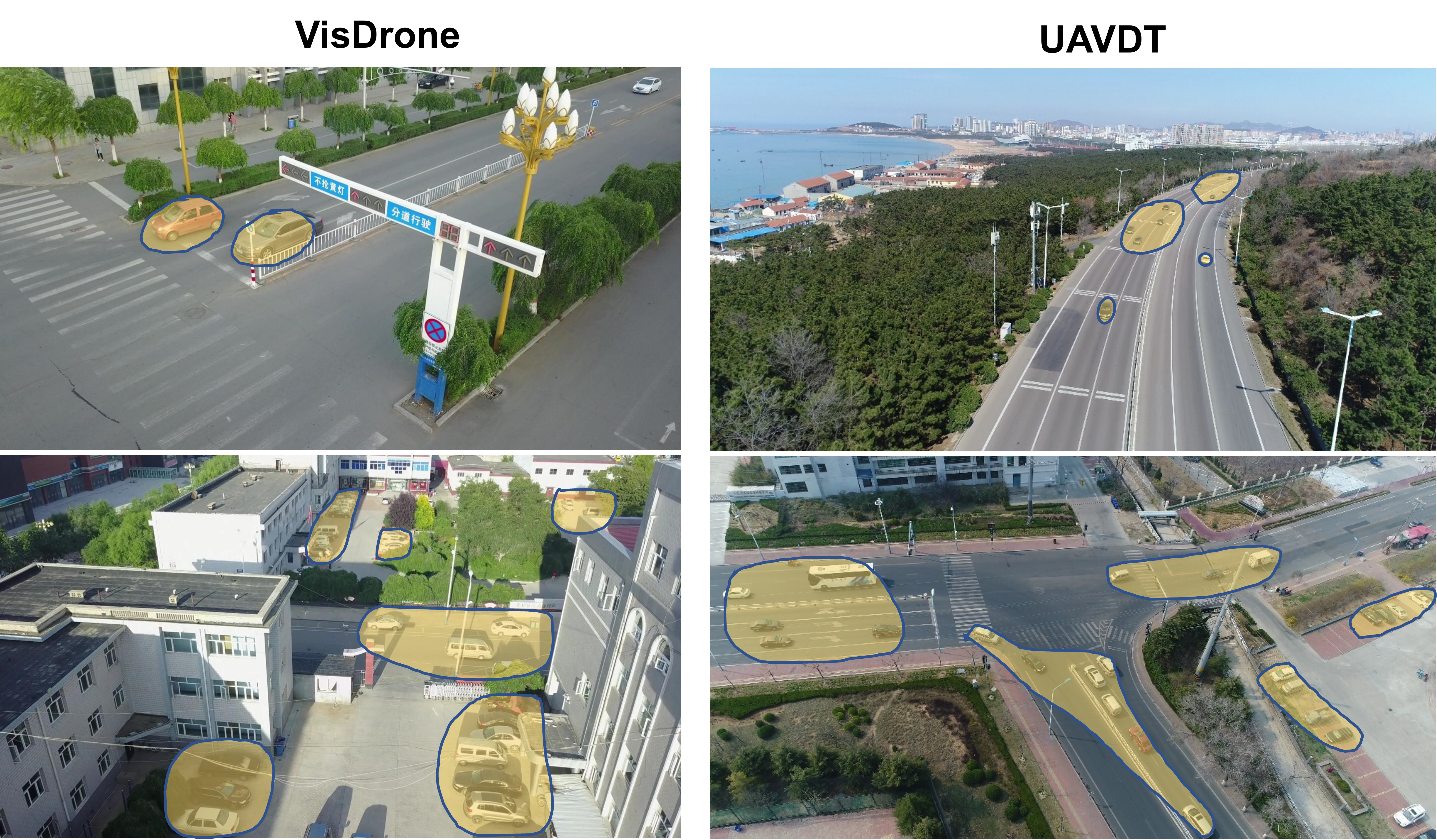}
        \caption{}
        \label{fig:short-b}
    \end{subfigure}
    \caption{(a) Comparison of foreground proportions on the COCO and drone imagery databases; and (b) visualization of foregrounds (highlighted in yellow) on samples from VisDrone and UAVDT.}
    \label{fig:challenge}
\end{figure}

To deal with the dilemma of balancing the accuracy and efficiency, a number of efforts are made, mainly on general object detection, which basically concentrate on reducing the complexity of the backbone networks~\cite{MobileNetV12017, ShufflenetV12018, DistillationDetection2017}. Despite some potential, these methods leave much room for improvement since they fail to take into account the heavy detection heads which are widely used by the state-of-the-art detectors~\cite{RetinaNet2017,ATSS2020,GFLv12020,ufpmpdet2022}. For instance, RetinaNet~\cite{RetinaNet2017} taking ResNet18~\cite{ResNet2016} as backbone with 512 input channels adopts a detection head that occupies 82.3\% of the overall GFLOPs. Recently, several methods have been presented to solve this problem, including network pruning~\cite{groupfisherprune2021, slimyolov32019} and structure redesigning~\cite{yolov42020, yolox2021}, and prove effective in accelerating inference. However, the former is criticized by the sharp performance drop when computations are greatly decreased, evidenced by the attempt on detection for UAVs \cite{slimyolov32019}, and the latter is primarily optimized for low-resolution input (\emph{e.g.} $640 \times 640$), making it not straightforward to adapt to high-resolution aerial images.

Sparse convolutions \cite{SECOND2018, SACT2017} show another promising alternative, which limit computations by only operating convolutions on sparsely sampled regions or channels via learnable masks. While theoretically attractive, their results are highly dependent on the selection of meaningful areas, because the focal region of the learned mask in sparse convolutions is prone to locate within foreground. Regarding drone images, the vast majority of objects are of small scales (as shown in Fig.~\ref{fig:challenge} (a)) and the scale of foreground areas varies along with flying altitudes and observing viewpoints (as shown in Fig.~\ref{fig:challenge} (b)), and this issue becomes even more prominent. An inadequate mask ratio enlarges the focal part and more unnecessary computations are consumed on background, which tends to simultaneously deteriorate efficiency and accuracy. On the contrary, an exaggerated one shrinks the focal part and incurs the difficulty in fully covering foreground and crucial context, thus leading to performance degradation. DynamicHead~\cite{DynamicHead2020} and QueryDet~\cite{QueryDet2022} indeed apply sparse convolutions to the detection head; unfortunately, their primary goal is to offset the increased computational cost when additional feature maps are jointly used for performance gain on general object detection. They both follow the traditional way in original sparse convolutions that set fixed mask ratios or focus on foreground only and are thus far from reaching the trade-off between accuracy and efficiency required by UAV detectors. Therefore, it is still an open question to leverage sparse convolutions to facilitate lightweight detection for UAVs.

In this paper, we propose a novel plug-and-play detection head optimization approach to efficient object detection on drone images, namely global context-enhanced adaptive sparse convolution (CEASC). Concretely, we first develop a context-enhanced sparse convolution (CESC) to capture global information and enhance focal features, which consists of a residual structure with a context-enhanced group normalization (CE-GN) layer. Since CE-GN specifically preserves a set of holistic features and applies their statistics for normalization, it compensates the loss of context caused by sparse convolutions and stabilizes the distribution of foreground areas, thus bypassing the sharp drop on accuracy. We then propose an adaptive multi-layer masking (AMM) scheme, and it separately estimates an optimal mask ratio by minimizing an elaborately designed loss at distinct levels of feature pyramid networks (FPN), balancing the detection accuracy and efficiency. It is worth noting that CESC and AMM can be easily extended to various detectors, indicating that CEASC is generally applicable to existing state-of-the-art object detectors for acceleration on drone imagery.


The contribution of our work lies in three-fold:

1) We propose a novel detection head optimization approach based on sparse convolutions, \emph{i.e.} CEASC, to efficient object detection for UAVs.

2) We introduce a context-enhanced sparse convolution layer and an adaptive multi-layer masking scheme to optimize the mask ratio, delivering an optimal balance between the detection accuracy and efficiency.

3) We extensively evaluate the proposed approach on two major public benchmarks of drone imagery by integrating CEASC to various state-of-the-art detectors (\eg RetinaNet and GFL V1), significantly reducing their computational costs while maintaining competitive accuracies. 

\section{Related Work}
\subsection{General Object Detection}

General object detection methods can be mainly divided into anchor-based detectors and anchor-free detectors depending on whether they use preset sliding windows or anchors to locate object proposals. In anchor-based detectors, the multi-stage detectors, including R-CNN~\cite{RCNN2014}, Faster-RCNN~\cite{Faster-RCNN2015}, Mask RCNN~\cite{MaskRCNN2017}, first generate proposal regions and subsequently classify and localize target objects within them. On the contrary, classification and localization of objects can be directly conducted in the whole feature on the one-stage detectors such as RetinaNet~\cite{RetinaNet2017} and GFL V1/V2~\cite{GFLv12020,GFLv22021}, which treat anchors as final bounding box targets. As for the anchor-free ones (\eg Centernet~\cite{centernet2019}, FCOS~\cite{FCOS2019} and FSAF~\cite{FSAF2019}), the anchors that incur heavy computational burden are replaced by efficient alternatives such as centerness constraints or object heatmaps. Although gains are consistently delivered, it is not so straightforward to adapt such methods to the case on UAVs.

\begin{figure*}[t]
\centering
\includegraphics[width=1\linewidth]{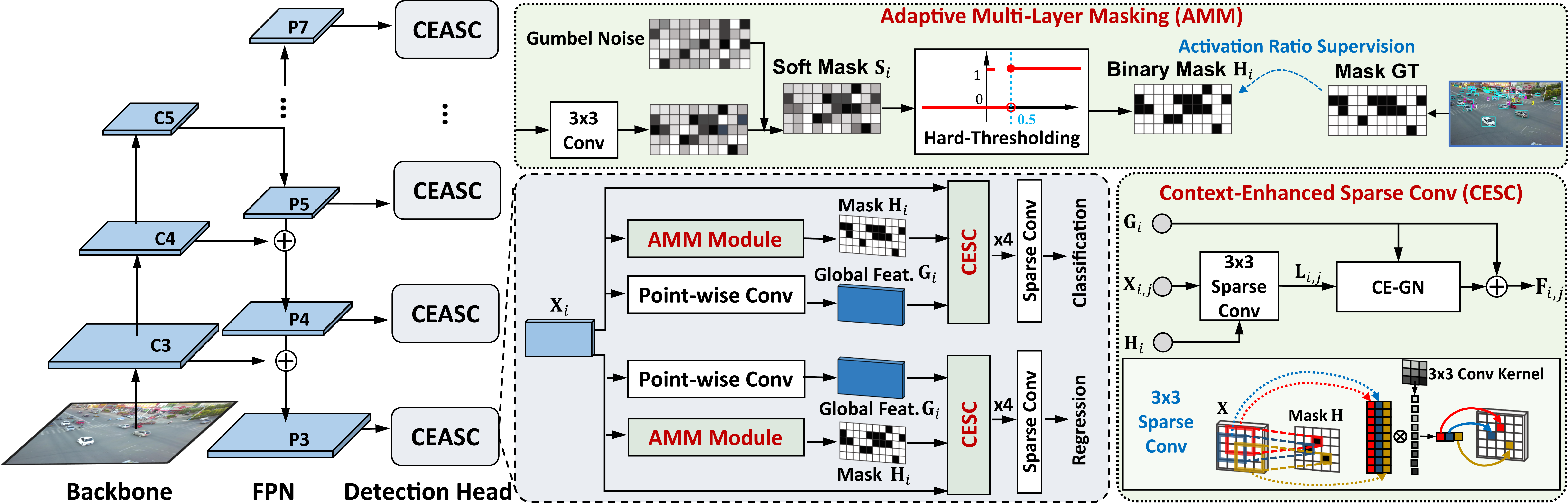}
\caption{Framework of CEASC. Given a base detector such as GFL V1, CEASC replaces the detection head by context-enhanced sparse convolution (CESC) in each FPN layer, via generating a mask feature $\mathbf{H}_{i}$ and a global feature $\mathbf{G}_{i}$ for context enhancement. The mask ratio of $\mathbf{H}_{i}$ is automatically optimized by the adaptive multi-layer masking (AMM) scheme, promoting both the accuracy and efficiency.} 
\label{fig:framework}
\vspace{0.1cm}
\end{figure*}
\subsection{Object Detection on Aerial Images}



For object detection on drone imagery, current studies usually follow a coarse-to-fine pipeline where a coarse detector is launched to locate large-scale instances and sub-regions that contain densely distributed small ones and a fine detector is further applied to those regions to find instances of small sizes. For example, ClusDet~\cite{ClusDet2019} employs a scale estimation network (ScaleNet) for better fine detection; DMNet~\cite{DMNet2020} optimizes region selection by conducting a density map guided connected crop generation; UFPMP-Det~\cite{ufpmpdet2022} merges sub-regions generated by a coarse detector into a unified image and designs the multi-proxy detection network to improve the detection accuracy of tiny objects; and Focus\&Detect~\cite{FocusAndDetect2022} makes use of the Gaussian mixture model to estimate focal regions and introduces incomplete box suppression to deal with overlapping focal areas. Despite of high accuracies achieved, these methods need to perform inference on one image for multiple times, which are not efficient, limiting their applications on the resource-constrained UAV platforms. 

\subsection{Lightweight Models for Object Detection}



Along with the advancement of deep learning, the complexity of object detection models has sharply increased, incurring heavy computational cost and slow inference speed. Several typical solutions are proposed in parallel to reduce computations for acceleration, including neural architecture search~\cite{efficientnet2019, nasfcos2020}, network pruning~\cite{networkslimming2017, groupfisherprune2021}, knowledge distillation~\cite{DistillationDetection2017, FGD2022} and lightweight model design~\cite{MobileNetV22018, thundernet2019}. Among them, lightweight model design is in the lead for detection on UAVs for its good potential in speed-accuracy trade-off.

Some methods focus on lightweight backbones, where MobileNet~\cite{MobileNetV12017,MobileNetV22018,MobileNetV32019} and ShuffleNet~\cite{ShufflenetV12018,ShufflenetV22018} are the representatives, which employ depth-wise separable convolutions and group convolutions, respectively. Some methods design lightweight detection heads, \emph{e.g.} in the YOLO series, YOLO v6~\cite{yolov62022} presents an efficient decoupled head, while YOLO v7~\cite{yolov72022} plans re-parameterized convolutions.


Sparse CNN has recently emerged as a promising way to accelerate inference by generating pixel-wise sample masks for convolutions. In particular, \cite{DynamicHead2020, QueryDet2022} have attempted to apply sparse convolutions to the detection head. ~\cite{DynamicHead2020} conducts a pixel-level combination of FPN features from different scales via spatial gates to reduce the computational cost. QueryDet~\cite{QueryDet2022} works on high-resolution images and utilizes the $P_{2}$ features from FPN to improve the accuracy on tiny objects, while a cascade sparse query structure is built and trained by the focal loss \cite{RetinaNet2017} for acceleration. Nevertheless, as these methods usually use a fixed mask ratio without capturing global context, they fail to handle severe fluctuations of foreground regions, leading to insufficiently optimized detection results on drone images. In contrast, our method adaptively adjusts the mask ratio with global feature captured to balance the efficiency and accuracy.

\section{Method}




As Fig.~\ref{fig:framework} shows, given a base detector, the entire CEASC network aims to optimize the detection head at different layers in FPN, by developing a context-enhanced sparse convolution (CESC), which integrates focal information with global context through a lightweight convolutional module as well as a context-enhanced group normalization (CE-GN) layer. An adaptive multi-layer masking (AMM) module is designed to enable the model adaptively generating masks with an adequate mask ratio, thus reaching a better balance in accuracy and efficiency. 


The details of the components aforementioned are described in Sec.~\ref{sec:CESC} and Sec.~\ref{sec:amm}.

\subsection{Context-Enhanced Sparse Convolution}\label{sec:CESC}
\subsubsection{Sparse Convolution}


Most existing detectors on drone images work with dense detection heads, convolving on the whole feature maps. Although fully exploring visual clues facilitates detecting tiny objects, the dense head requires much more computations, which is not applicable to the resource-constrained UAV platform. In the mean time, the foreground area only occupies a small part of a frame acquired by a drone as shown in Fig.~\ref{fig:challenge}, indicating that the dense head conducts a lot of computational operations on background, which contains much less useful information for object detection. This observation reveals the potential to accelerate the detection head by only computing on the foreground area. 


Sparse convolution (SC) \cite{SECOND2018, SACT2017} have recently been proposed, which learn to operate on foreground areas by employing a sparse mask and prove effective in speeding up the inference phase on a variety of vision tasks. Inspired by them, we construct our network based on SC.


Specifically, given a feature map $\mathbf{X}_{i} \in \mathbb{R}^{B \times C \times H \times W}$ from the $i$-th layer of FPN, SC adopts a mask network consisting of a shared kernel $\mathbf{W}_{mask} \in \mathbb{R}^{C \times 1 \times 3 \times 3}$, where $B$, $C$, $H$, $W$ refers to the batch size, channel size, height and width, respectively. Convolving on $\mathbf{X}_{i}$ based on $\mathbf{W}_{mask}$ generates a soft feature $\mathbf{S}_{i} \in \mathbb{R}^{B \times 1 \times H \times W}$, which is further turned to a mask matrix $\mathbf{H}_{i} \in \{0,1\}^{B \times 1 \times H \times W}$ by using the Gumbel-Softmax trick \cite{DynamicConvolution2020} formulated as below:
\begin{equation}
\mathbf{H}_{i} =
\begin{cases}
\sigma\Bigl(\frac{\mathbf{S}_{i} + g_{1} - g_{2}}{\tau}\Bigr) > 0.5, & \text{ For training } \\
\mathbf{S}_{i} > 0, &  \text{ For inference }
\end{cases}
\label{eq:hard-mask}
\end{equation}
where $g_{1}, g_{2} \in \mathbb{R}^{B \times 1 \times H \times W}$ denote two random gumbel noises, $\sigma$ refers to the sigmoid function, and $\tau$ is the corresponding temperature parameter in Gumbel-Softmax.


According to Eq. \eqref{eq:hard-mask}, only the area with the mask value 1 involves in convolutions during inference, thus reducing the overall computational cost. The sparsity of $\mathbf{H}_{i}$ is controlled by a mask ratio $r\in [0,1]$, which is often set larger than 0.9 by hand in existing studies. Since the base detector (here we take GFL V1 as an example) has a classification head and a regression head in the detection framework, we separately introduce a mask network for each head considering that they often focus on different areas. Each detection head adopts four Convolution-GN-ReLU layers and a single convolution layer to make prediction, where we replace the conventional convolution layers with the SC ones.

\subsubsection{Context Enhancement}\label{sec:CE}




As claimed in \cite{FGD2022}, contextual clues (\eg background surrounding target objects) benefit object detection; however, SC performs convolutions only on foreground and abandons background with useful information, which probably undermines the overall accuracy, especially in the presence of tiny objects prevailing in drone images. To tackle with this problem, \cite{SpatiallyAdaptiveInference2020} attempts to recover surrounding context by interpolation, but it is not reliable as the focal and background areas exhibit large discrepancy. In this work, we propose a lightweight CESC module, jointly making use of focal information and global context for enhancement and simultaneously boosting the stability of subsequent computations. As displayed in Fig.~\ref{fig:framework}, we apply a point-wise convolution to the feature map $\mathbf{X}_{i}$, generating the global contextual feature $\mathbf{G}_{i}$. Since only a few elements in $\mathbf{X}_{i}$ are processed by SC, $\mathbf{G}_{i}$ tends to become stable after multiple rounds of SC without taking much extra computational cost.


As an important part of SC, we embed the global contextual information $\mathbf{G}_{i}$ into the SparseConvolution-GN-ReLU layers, which takes the feature map $\mathbf{X}_{i,j}$, the mask $\mathbf{H}_{i}$, and the global feature $\mathbf{G}_{i}$ as inputs, where $j$ indicates the $j$-th SparseConvolution-GN-ReLU layer. Instead of using the activated elements to compute the statistics for group normalization as in conventional SC, we adopt the mean value and standard deviation of $\mathbf{G}_{i}$ for normalization, aiming to compensate the missing context. Supposing that $\mathbf{L}_{i,j}$ is the output feature map after applying SC on $\mathbf{X}_{i,j}$, the context-enhanced feature $\mathbf{F}_{i,j}$ is obtained by CE-GN as below

\begin{table*}[!t]
    \centering
    \normalsize
    \begin{tabular}{c|c|ccc|cccc|cc}
    \hline
    Base Detector & Method & mAP & $\text{AP}_{50}$ & $\text{AP}_{75}$ & $\text{AR}_{1}$ & $\text{AR}_{10}$ & $\text{AR}_{100}$ & $\text{AR}_{500}$ & GFLOPs & FPS \\ \hline
    \multirow{2}*{GFL V1~\cite{GFLv12020}}        & Baseline & 28.4 & 50.0 & 27.8 & 0.62 & 6.36 & 35.6 & 44.9 & 524.95 & 13.46 \\
                                 & \textbf{Ours (CEASC)}  & \textbf{28.7} & \textbf{50.7} & \textbf{28.4} & \textbf{0.65} & \textbf{6.56} & 35.6 & \textbf{45.0} & \textbf{150.18} & \textbf{21.55} \\ \hline
    \multirow{2}*{RetinaNet~\cite{RetinaNet2017}}     & Baseline & \textbf{21.8} & 39.3 & \textbf{21.1} & 0.54 & 5.82 & \textbf{29.1} & \textbf{35.3} & 529.81 & 13.41 \\
                                 & \textbf{Ours (CEASC)}  & 21.6 & \textbf{39.6} & 20.6 & \textbf{0.59} & 5.82 & 28.9 & 34.7 & \textbf{157.41} & \textbf{20.10} \\ \hline
     \multirow{2}*{Faster-RCNN~\cite{Faster-RCNN2015}}  & Baseline & \textbf{24.8} & \textbf{43.6} & \textbf{25.0} & 0.64 & \textbf{5.97} & \textbf{33.0} & 34.9 & 322.25 & 18.17 \\
                                 & \textbf{Ours (CEASC)}  & 24.6 & 43.4 & 24.7 & 0.64 & 5.91 & 32.8 & \textbf{35.1} & \textbf{132.91} & \textbf{21.71} \\ \hline
     \multirow{2}*{FSAF~\cite{FSAF2019}}         & Baseline & \textbf{26.3} & \textbf{50.3} & \textbf{23.7} & 0.53 & 5.25 & \textbf{32.5} & \textbf{43.5} & 518.25 & 14.06 \\
                                 & \textbf{Ours (CEASC)}  & 25.0 & 48.9 & 22.0 & \textbf{0.56} & \textbf{5.65} & 31.1 & 41.5 & \textbf{153.92} & \textbf{19.43} \\ \hline
    
    \end{tabular}
    \caption{Comparison of AP/AR (\%) and GFLOPs/FPS on VisDrone by using our approach with various base detectors.}
    \vspace{0.1cm}
    \label{tab:sota_diff_basedet}
\end{table*}

\begin{equation}
\mathbf{F}_{i,j} = w \times \frac{\mathbf{L}_{i,j} - mean[\mathbf{G}_{i}]}{{std}[\mathbf{G}_{i}]} + b
\label{eq:ce-gn}
\end{equation}
where $mean[\cdot]$ and $std[\cdot]$ denote the mean and standard deviation, respectively, and $w$ and $b$ are learnable parameters.

To further mitigate the information loss in SC and make the training process more stable, we additionally maintain the normal dense convolution besides the sparse one during training, generating a feature map $\mathbf{C}_{i,j}$ convolved on the full input feature map. We then employ $\mathbf{C}_{i,j}$ to enhance the sparse feature map $\mathbf{F}_{i,j}$ by optimizing the MSE loss as: 
\begin{equation}
\mathcal{L}_{norm} = \frac{1}{4L}\sum_{i=1}^{L}\sum_{j=1}^{4}\|\mathbf{C}_{i,j} \times \mathbf{H}_{i}\ -\ \mathbf{F}_{i,j}\|^{2},\label{eq:norm_loss}
\end{equation}
where $L$ is the amount of layers in FPN.


We finally adopt a residual structure before the activation layer by adding $\mathbf{G}_{i}$ to $\mathbf{F}_{i,j}$, \ie $\mathbf{F}_{i,j}:=\mathbf{F}_{i,j}+\mathbf{G}_{i}$, which strengthens context preservation. The complete architecture of the CESC module and the CE-GN layer are displayed in Fig.~\ref{fig:framework}.

\subsection{Adaptive Multi-layer Masking}
\label{sec:amm}



Without any extra constraint, the sparse detector tends to generate the mask with a large activation ratio (or a small mask ratio) for a higher accuracy, thus increasing the overall computational cost. To deal with this issue, most existing attempts use a fixed activation ratio. However, as the foreground of aerial images exhibits severe fluctuations, a fixed ratio is prone to incur either significant increase in computation or decrease in accuracy due to insufficient coverage over foreground areas. For the trade-off between accuracy and efficiency, we propose the AMM scheme to adaptively control the activation ratio (or reversely the mask ratio). 



Specifically, AMM firstly estimates an optimal mask ratio based on the ground-truth label. By leveraging the label assignment technique, for the $i$-th FPN layer, we obtain the ground-truth classification results $\mathcal{C}_{i} \in \mathbb{R}^{h_{i} \times w_{i} \times c}$, where $c$ represents the number of categories including the background; $h_{i}$ and $w_{i}$ indicate the height and width of the feature map, respectively. The optimal activation ratio $\mathcal{P}_{i}$ in the $i$-th FPN layer is estimated as


\begin{equation}
\mathcal{P}_{i} = \frac{Pos(\mathcal{C}_{i})}{{Numel(\mathcal{C}_{i})}},
\label{eq:p}
\end{equation}
where $Pos(\mathcal{C}_{i})$ and ${Numel(\mathcal{C}_{i})}$ indicate the number of pixels belonging to the positive (foreground) instances and that of all pixels, respectively. 

To guide the network adaptively generating a mask with an adequate mask ratio, we employ the following loss  
\begin{equation}
\mathcal{L}_{amm} = \frac{1}{L}\sum_{i}\Bigl(\frac{Pos(\mathbf{H}_{i})}{Numel(\mathbf{H}_{i})} - \mathcal{P}_{i}\Bigr)^{2},
\label{eq:loss_cost}
\end{equation}
where $\frac{Pos(\mathbf{H}_{i})}{Numel(\mathbf{H}_{i})}$ indicates the activation ratio of the mask $\mathbf{H}_{i}$. By minimizing $\mathcal{L}_{amm}$, $\mathbf{H}_{i}$ is forced to abide by the same activation ratio as the ground-truth foreground ratio $\mathcal{P}_{i}$, thus facilitating the generation of adequate mask ratios.

By adding the conventional detection loss $\mathcal{L}_{det}$, we formulate the overall training loss as follows:
\begin{equation}
\mathcal{L} = \mathcal{L}_{det} + \alpha \times \mathcal{L}_{norm} + \beta \times \mathcal{L}_{amm}\label{eq:overall_loss},
\end{equation}
where $\alpha$, $\beta$ are hyper-parameters balancing the importance of $\mathcal{L}_{norm}$ and $\mathcal{L}_{amm}$.

\section{Experiments}
We evaluate the effectiveness of CEASC by comparing it to the state-of-the-art lightweight approaches and conducting comprehensive ablation studies.

\subsection{Datasets and Metrics}


We adopt two major benchmarks for evaluation in drone-based object detection, \ie VisDrone~\cite{VisDrone2018} and UAVDT~\cite{UAVDT2018}. VisDrone consists of 7,019 high-resolution (2,000$\times$1,500) aerial images belonging to 10 categories. Following previous work \cite{ClusDet2019, QueryDet2022}, we use 6,471 images for training and 548 images for testing. UAVDT contains 23,258 training images and 15,069 testing images with a resolution of 1,024$\times$540 from 3 classes. 


We employ the mean Average Precision (mAP), Average Precision (AP) and Average Recall (AR) as the evaluation metrics on accuracy, as well as GFLOPs and FPS as the ones on efficiency.

\subsection{Implementation Details}

We implement our network based on PyTorch~\cite{pytorch2019} and MMDetection~\cite{mmdetection2019}. On VisDrone, all models are trained for 15 epochs with the SGD optimizer, and the learning rate is initialized as 0.01 with a linear warm-up and decreased by 10 times after 11 and 14 epochs. On UAVDT, we train models for 6 epochs with an initial learning rate at 0.01, decreased by 10 times after 4 and 5 epochs. The trade-off hyper-parameters $\alpha$ and $\beta$ in Eq. \eqref{eq:overall_loss} are set to 1 and 10, respectively, and the temperature parameter $\tau$ in Gumbel Softmax is fixed as 1. We make use of GFL V1 as the base detector and ResNet18 as the backbone with 512 feature channels by default. The input image sizes are set to 1,333$\times$800 and 1,024$\times$540 on VisDrone and UAVDT, respectively. All experiments are conducted on two NVIDIA RTX 2080Ti GPUs, except that the inference speed is test on a single RTX 2080Ti GPU. 


\subsection{Evaluation on Different Detectors}
\label{sec:performance-sota-detector}


It is worth noting that the proposed CEASC network is plug-and-play. To validate its effect in a wide range of base detectors, we report the performance by combining CEASC with four prevailing base detectors: GFL V1 \cite{GFLv12020}, RetinaNet \cite{RetinaNet2017}, Faster-RCNN \cite{Faster-RCNN2015} and FSAF \cite{FSAF2019}. As shown in Table \ref{tab:sota_diff_basedet}, by integrating CEASC, the GFLOPs of all the base detectors are reduced by at least 60\%, and the FPS is promoted by 20\%$\sim$60\% with slight fluctuations in mAP, indicating its effectiveness and generalizability in accelerating detectors without sacrificing their accuracies.

\begin{table}[!t]
    \centering
        
    \normalsize
    \resizebox{\linewidth}{!}{
    \begin{tabular}{cc|ccc|cc}
    \hline
    CESC & AMM & mAP & $\text{AP}_{50}$ & $\text{AP}_{75}$ & GFLOPs & FPS \\ \hline
      &  & 28.4 & 50.0 & 27.8 & 524.95 & 13.46 \\
    \checkmark & & 28.6 & 50.6 & 28.2 & 158.23 & 19.26 \\
    \checkmark  & \checkmark & \textbf{28.7} & \textbf{50.7} & \textbf{28.4} & \textbf{150.18} & \textbf{21.55}\\
     \hline
    \end{tabular}
    }
    \caption{Ablation on CESC and AMM with GFL V1 as the base detector on VisDrone.}
    \vspace{0.1cm}
    \label{tab:ab_study}
\end{table}

\begin{figure}[!t]
\centering
\includegraphics[width=1\linewidth]{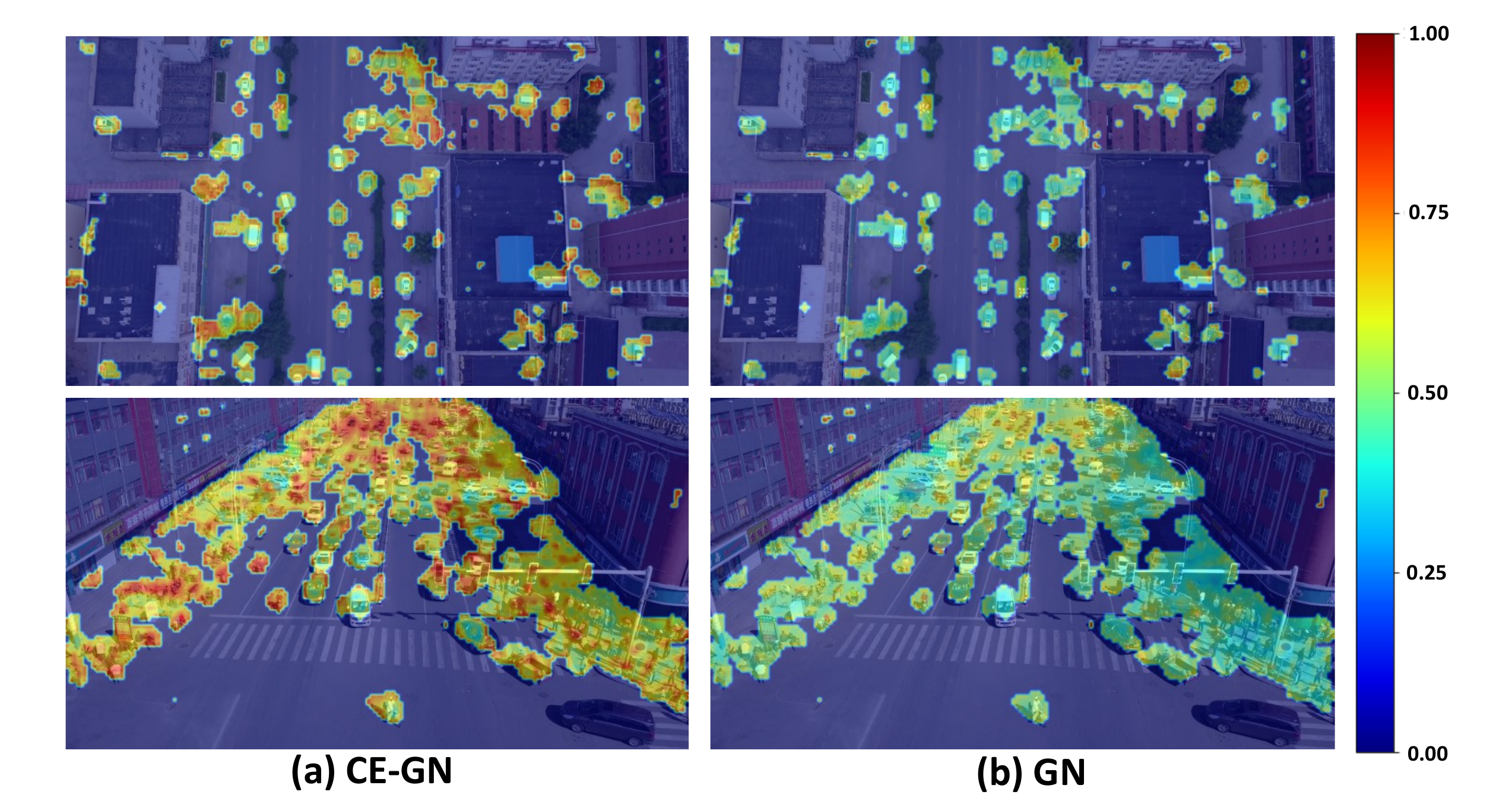}
\caption{Visualization on correlation between features generated by dense convolutions and those by sparse convolutions using distinct normalization schemes on VisDrone, (a) and (b) use CE-GN and GN on sparse convolutions, respectively.} 
\label{fig:ce-gn}
\vspace{0.1cm}
\end{figure}

\subsection{Ablation Study}\label{sec:ab_study}
We validate the main components of CEASC, where we also adopt GFL V1 as the base detector in all the ablation studies.  


\subsubsection{On CESC and AMM} 

As Table \ref{tab:ab_study} reports, by employing the CESC component, the base detector saves about 70\% of GFLOPs and runs 1.43 times faster without any drop in accuracy, since SC reduces the complexity and the CE-GN layer together with the residual structure compensates the loss of context. By adopting the dynamic mask ratio to obtain a compact foreground coverage, the AMM component further increases the accuracy and accelerates the inference speed by 11.9\% while saving 5.1\% of GFLOPs. Note that the training process of GFL V1 becomes extremely unstable when directly applying SC without CESC, and we thus do not provide the result by individually evaluating AMM on GFL V1. 

\begin{table}[!t]
    \centering
        
    \resizebox{\linewidth}{!}{
    \begin{tabular}{cccc|ccc|cc}
    \hline
     SC & Res. & CE-GN & $\mathcal{L}_{norm}$ & mAP & $\text{AP}_{50}$ & $\text{AP}_{75}$ & GFLOPs & FPS \\ \hline
    & &  &  & 28.4 & 50.0 & 27.8 & 524.95 & 13.46 \\
    \checkmark & \checkmark &  &  & 26.1 & 47.2 & 25.3 & 151.66 & 21.51 \\
    \checkmark & \checkmark & \checkmark &  & 28.5 & 50.5 & 28.3 & 155.90 & 19.91 \\
    \checkmark & \checkmark & \checkmark & \checkmark & \textbf{28.7} & \textbf{50.7} & \textbf{28.4} & \textbf{150.18} & \textbf{21.55} \\ \hline
    \end{tabular}}
    \caption{Ablation on detailed designs in CESC with GFL V1 on VisDrone.}
    \label{tab:ab_cesc}
\end{table}

\begin{table}[!t]
    \centering
        
    \resizebox{\linewidth}{!}{
    \begin{tabular}{c|ccc|cc}
    \hline
    Method & mAP & $\text{AP}_{50}$ & $\text{AP}_{75}$ & GFLOPs & FPS \\ \hline
    Dense Conv. & 28.4 & 50.0 & 27.8 & 524.95 & 13.46 \\
    w/o Normalization & 26.1 & 47.2 & 25.3 & 151.66 & 21.51 \\
    GN~\cite{GN2018} & 28.0 & 49.9 & 27.7 & 154.49 & 18.82 \\
    BN~\cite{BatchNorm2015} & 26.1 & 47.0 & 25.4 & 150.81 & 19.55 \\
    IN~\cite{InstanceNorm2016} & 27.9 & 49.7 & 27.6 & 160.91 & 19.30 \\
    CE-GN (\textbf{Ours}) & \textbf{28.7} & \textbf{50.7} & \textbf{28.4} & \textbf{150.18} & \textbf{21.55} \\ \hline
    \end{tabular}}
    \caption{Ablation on CE-GN with GFL V1 on VisDrone.}
    \label{tab:ab_cegn}
\end{table}

\begin{table}[!t]
    \centering
       
    \resizebox{\linewidth}{!}{
    \begin{tabular}{c|ccc|cc}
    \hline
    Method & mAP & $\text{AP}_{50}$ & $\text{AP}_{75}$ & GFLOPs & FPS \\ \hline
    $3\times3$ convolution & 28.5 & 50.1 & 28.1 & 262.38 & 17.12 \\
    GhostModule~\cite{ghostnet2020} & 28.3 & 50.1 & 27.8 & 194.66 & 19.22 \\
    CBAM~\cite{cbam2018} & 28.4 & 50.3 & 27.8 & \textbf{148.08} & 16.20 \\
    Criss-Cross Attn.~\cite{ccnet2019} & 28.4 & 50.3 & 27.8 & 159.27 & 15.40 \\
    Point-wise (\textbf{Ours}) & \textbf{28.7} & \textbf{50.7} & \textbf{28.4} & 150.18 & \textbf{21.55}\\ \hline
    \end{tabular}}
     \caption{Comparison of distinct methods to encode global context with GFL V1 on VisDrone.}
    \label{tab:ab_global_info}
\end{table}

\begin{figure*}[!t]
\centering
\includegraphics[width=1\linewidth]{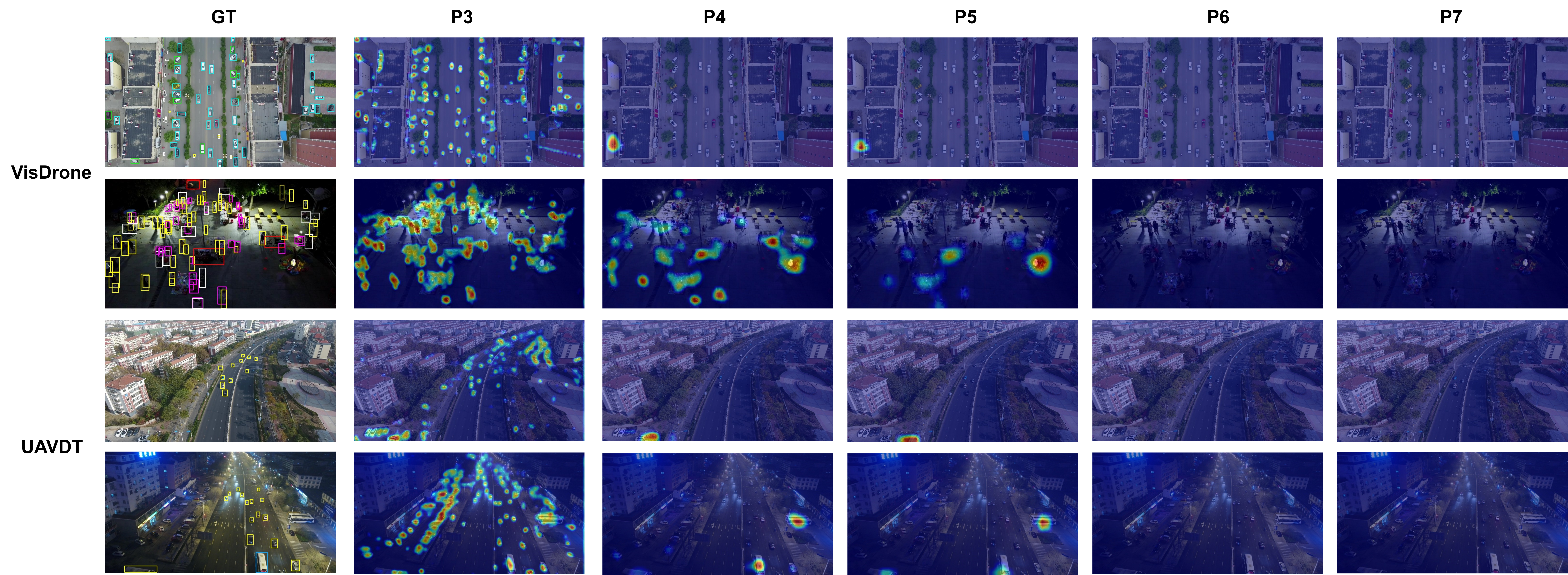}
\caption{Visualization of the dynamic masks estimated by AMM for different layers (from `P3' to `P7') in FPN of GFL V1. Highlighted areas are activated for computation.}
\label{fig:vis}
\vspace{0.1cm}
\end{figure*}

\begin{table}[!t]
    \centering
    \small  
    \begin{tabular}{c|ccc|cc}
    \hline
    Method & mAP & $\text{AP}_{50}$ & $\text{AP}_{75}$ & GFLOPs & FPS \\ \hline
    Global & 28.4 & 50.2 & 28.1 & 162.53 & 19.84 \\
    Layer-wise & \textbf{28.7} & \textbf{50.7} & \textbf{28.4} & \textbf{150.18} & \textbf{21.55} \\ \hline
    \end{tabular}
    \caption{Comparison of estimating the mask ratio in different ways by AMM on VisDrone.}
    \label{tab:layerwise}
\end{table}

\begin{figure}[!tpb]
\centering
\begin{subfigure}{\linewidth}
        \centering
        \includegraphics[width=\linewidth]{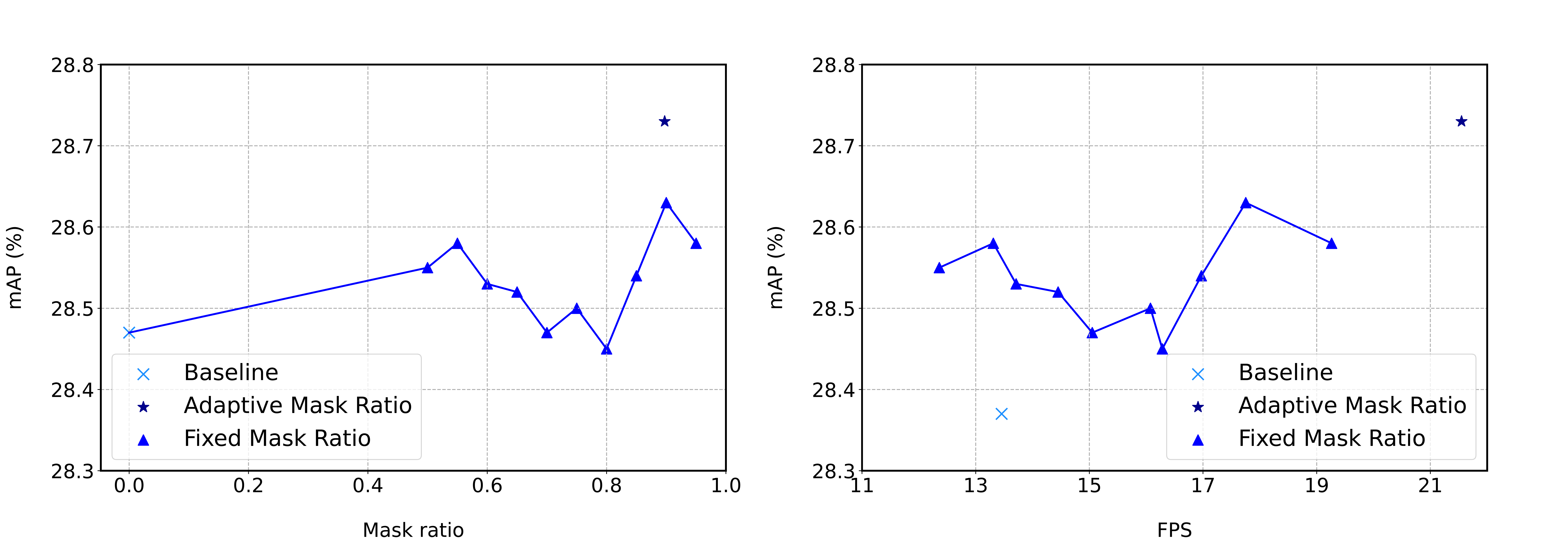}
        \caption{VisDrone}
        \label{fig:short-res-a}
    \end{subfigure}
    \hfill
    \begin{subfigure}{\linewidth}
        \centering
        \includegraphics[width=\linewidth]{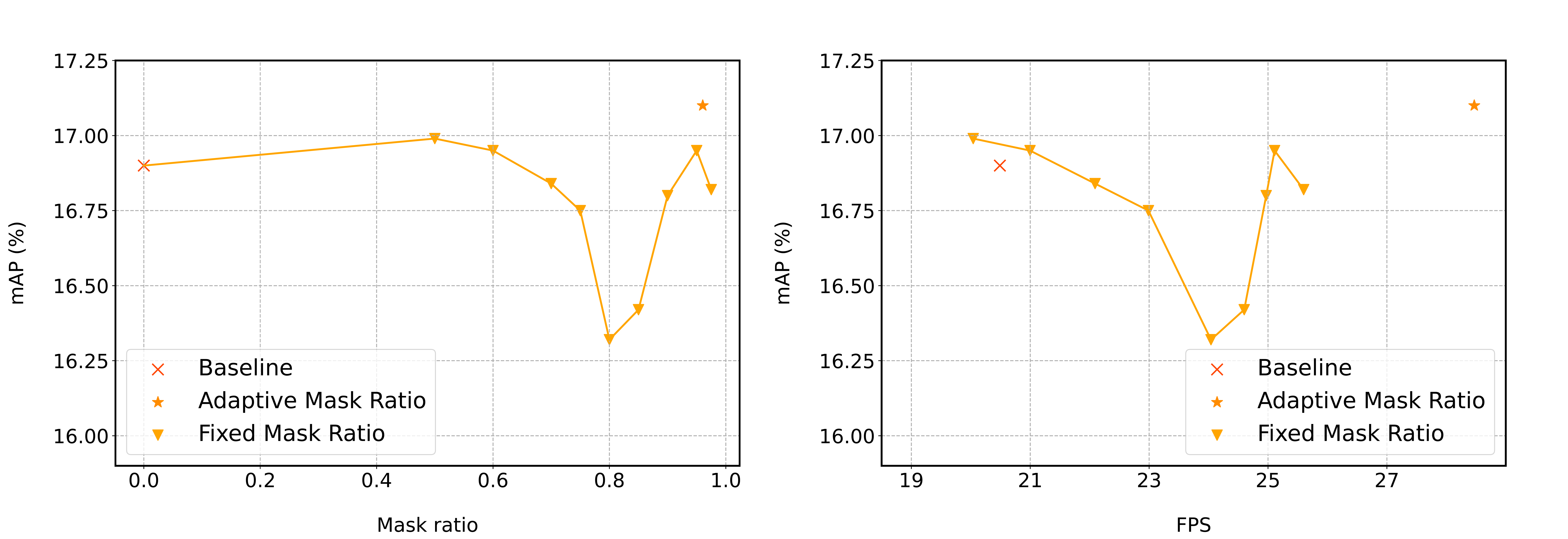}
        \caption{UAVDT}
        \label{fig:short-res-b}
    \end{subfigure}
\caption{Comparison of the fixed mask ratio and the dynamic one estimated by AMM.}\label{fig:ab_amm}
\vspace{0.2cm}
\end{figure}


\subsubsection{On Detailed Designs in CESC} We separately evaluate the effect of the residual structure (Res. for short), CE-GN and the normalization loss $\mathcal{L}_{norm}$ in Eq.~\eqref{eq:norm_loss} on the performance of CESC. Recall that directly applying SC to GFL V1 makes the training process unstable. As summarized in Table~\ref{tab:ab_cesc}, when employing the residual structure, GFL V1 with SC turns to be stable and requires much less GFLOPs, but the mAP sharply drops due to the loss of context. By adding the context information via CE-GN, the accuracy is significantly promoted with a slight increase in GFLOPs. $\mathcal{L}_{norm}$ further boosts the accuracy and efficiency, since it implicitly strengthens the sparsity of the features. 

We further evaluate the performance of CE-GN by comparing it to the counterparts including the one without using normalization as in QueryDet~\cite{QueryDet2022}, GroupNorm (GN)~\cite{GN2018} as in DynamicHead~\cite{DynamicHead2020}, BatchNorm (BN)~\cite{BatchNorm2015} and InstanceNorm (IN)~\cite{InstanceNorm2016}. We also report the results by the original GFL V1 detector denoted as `Dense Conv.'. As displayed in Table~\ref{tab:ab_cegn}, CE-GN substantially promotes the accuracy of the model without normalization by 2.6\%. In comparison to the other normalization schemes, CE-GN achieves the best accuracy, 0.7\%, 2,6\% and 0.8\% higher than GN, BN and IN, respectively. It is worth noting that CE-GN performs the best in efficiency in regards of GFLOPs and FPS as well. To highlight the advantages of CE-GN, we visualize the cosine similarities between the features generated by dense convolutions and sparse convolutions, where CE-GN and GN are separately utilized to normalize SC. As Fig.~\ref{fig:ce-gn} illustrates, the features using CE-GN exhibit higher correlations than those using GN, showing the superiority of CE-GN in enhancing global context for SC.

\begin{table}[!t]
    \centering
    \resizebox{\linewidth}{!}{
    \begin{tabular}{c|ccc|cc}
    \hline
    Method & mAP & $\text{AP}_{50}$ & $\text{AP}_{75}$ & GFLOPs & FPS \\ \hline
    with P3 & 26.9 & 48.6 & 26.3 & \textbf{143.03} & \textbf{27.78} \\
    with P3-P4 & 28.5 & 50.5 & 28.1 & 149.09 & 24.60 \\
    with P3-P5 & \textbf{28.7} & \textbf{50.7} & \textbf{28.4} & 150.01 & 21.79 \\
    \textbf{Ours} (with P3-P7) & \textbf{28.7} & \textbf{50.7} & \textbf{28.4} & 150.18 & 21.55\\ \hline
    \end{tabular}}
    \caption{Ablation on FPN with GFL-V1 on VisDrone.}
    \label{tab:ab_fpn}
\end{table}

\begin{table*}[!t]
    \centering 
    \normalsize
    \begin{tabular}{c|c|c|ccc|ccc}
    \hline
     Base Detector & Method & Backbone & mAP & $\text{AP}_{50}$ & $\text{AP}_{75}$ & GFLOPs & FPS \\ \hline
    \multirow{4}*{GFL V1~\cite{GFLv12020}} & Baseline     & ResNet18 & 28.4 & 50.0 & 27.8 & 524.95 & 13.46 \\
     & MobileNet V2~\cite{MobileNetV22018}                          & MobileNet V2 & 28.5 & 50.2 & 28.1 & 491.47 & 13.63 \\ 
     & ShuffleNet V2~\cite{ShufflenetV22018}                         & ShuffleNet V2 & 26.2 & 46.6 & 25.7 & 488.94 & 13.92 \\  
     & \textbf{Ours (CEASC)}                              & ResNet18 & \textbf{28.7} & \textbf{50.7} & \textbf{28.4} & \textbf{150.18} & \textbf{21.55} \\ \hline
    \multirow{3}*{RetinaNet~\cite{RetinaNet2017}} & Baseline  & ResNet50 & 20.2 & \textbf{36.9} & 19.5 & 586.77 & 10.27 \\
     & QueryDet~\cite{QueryDet2022}                          & ResNet50 & 19.6 & 35.7 & 19.0 & - & 10.65 \\
     & QueryDet-CSQ~\cite{QueryDet2022}                      & ResNet50 & 19.3 & 35.0 & 18.9 & - & 11.71 \\
     & \textbf{Ours (CEASC)}                             & ResNet50 & \textbf{20.8} & 35.0 & \textbf{27.7} & \textbf{201.96} & \textbf{14.27} \\ \hline
    \end{tabular}
    \caption{Comparison of mAP/AP (\%) and GFLOPs/FPS with the state-of-the-art approaches on VisDrone. `-' indicates that the result is not reported or not publicly available.}
    \vspace{0.1cm}
    \label{tab:sota_on_visdrone}
    
\end{table*}

\begin{table}[!t]
    \centering
    
    \normalsize
    \resizebox{\linewidth}{!}{
    \begin{tabular}{c|ccc|cc}
    \hline
    Method & mAP & $\text{AP}_{50}$ & $\text{AP}_{75}$ & GFLOPs & FPS \\ 
    \hline
    Baseline & 16.9 & 29.5 & \textbf{17.9} & 271.66 & 20.49 \\
    \textbf{Ours (CEASC)}  & \textbf{17.1} & \textbf{30.9} & 17.8 & \textbf{64.12} & \textbf{28.47} \\ \hline
    \end{tabular}}
    \caption{Comparison of mAP/AP (\%) and GFLOPs/FPS with GFL V1 on UAVDT.}
    \label{tab:sota_uavdt}
\end{table}


To encode global context, we utilize the point-wise convolution, and make comparison to existing techniques including the plain $3\times 3$ convolution, GhostModule~\cite{ghostnet2020}, and several attention-based methods such as CBAM~\cite{cbam2018} and Criss-Cross Attention~\cite{ccnet2019}. As summarized in Table ~\ref{tab:ab_global_info}, the point-wise convolution outperforms the counterparts in detection accuracy. Meanwhile, it reaches the lowest GFLOPs in the convolution-based approaches and achieves the highest FPS among all the methods, clearly demonstrating its advantage in balancing the accuracy and efficiency.

\subsubsection{On Detailed Analysis of AMM} We compare the AMM module with a fixed mask ratio ranging from 0.50 to 0.95 on VisDrone and from 0.50 to 0.975 on UAVDT, respectively. As Fig.~\ref{fig:ab_amm} shows, more features are involved in convolution when reducing the mask ratio, resulting in higher computational cost and lower FPS. In the mean time, we can see that the detection accuracy is sensitive to the mask ratio, which is not consistently improved as the ratio increases. Moreover, the optimal fixed mask ratio varies on different datasets, \eg 0.9 on VisDrone and 0.95 on UAVDT in regards of mAP. In contrast, AMM adaptively determines an appropriate mask ratio, with which the base detector reaches the best accuracy and the highest inference speed, demonstrating its necessity.

Note that AMM separately computes the mask ratio for different layers in a ``Layer-wise'' way. We compare it to a ``Global'' version, which estimates a global mask ratio for all layers. As demonstrated in Table \ref{tab:layerwise}, the ``Layer-wise'' method clearly performs better than the ``Global'' one in terms of mAP and FPS. The reason lies in that the optimal mask ratio varies in different layers of FPN as displayed in Fig.~\ref{fig:vis}, and the ``Layer-wise'' method estimates the mask ratio more precisely than the ``Global'' one, thus promoting both the accuracy and efficiency. We also evaluate its effect at different FPN layers in Table~\ref{tab:ab_fpn}. With less FPN layers, GFLOPs and FPS are improved. Abandoning P6-P7 does not affect much as they are less informative. Removing P4 incurs a sharp drop in mAP, indicating that P4 is crucial, which is consistent with the visualization.

\subsection{Comparison to SOTA}

We compare our network with the state-of-the-art ones: 1) the lightweight methods including MobileNet V2~\cite{MobileNetV12017} and ShuffleNet V2~\cite{ShufflenetV12018}; 2) the detection head optimization methods for drone imagery including QueryDet~\cite{QueryDet2022} and its acceleration part QueryDet-CSQ~\cite{QueryDet2022}. Since GFL V1 \cite{GFLv12020} with ResNet18 as the backbone is widely used and proves effective in drone-based object detection, we select it as the base detector, and denote the original version as the `Baseline' method. We also report the result by using RetinaNet~\cite{RetinaNet2017} with ResNet50 as the backbone, since it is used as the base detector in QueryDet and QueryDet-CSQ. Note that the same data augmentation technique used in QueryDet is adopted in our implementation for fair comparison.


As summarized in Table \ref{tab:sota_on_visdrone}, CEASC remarkably reduces the GFLOPs of the base detectors (GFL V1 and RetinaNet), reaching a slightly higher mAP in the mean time. For instance, CEASC decreases the GFLOPS of the Baseline GFL V1 by 71.4\% and achieves 60\% speedup in terms of FPS during inference, with a 0.3\% improvement in mAP. Since the lightweight models, \ie MobileNet V2 and ShuffleNet V2, quest for efficiency by simplifying the network structures, their mAPs are lower than ours. Moreover, they apply dense detection heads, thus requiring much more GFLOPs. Though QueryDet-CSQ considers to optimize the detection head by the CSQ module with sparse convolutions, it only concentrates on small objects and ignores the loss of contextual information. Besides, QueryDet introduces an extra heavy query head to promote performance, which inevitably incurs more computational cost. In contrast, CEASC newly develops the context-enhanced sparse convolution module and designs an adaptive multi-layer masking scheme, thus clearly outperforming QueryDet and QueryDet-CSQ, both in accuracy and efficiency. 


We also evaluate CEASC on UAVDT. As reported in Table \ref{tab:sota_uavdt}, our method reduces the GFLOPs by 76.3\% and boosts the inference speed by 38.9\% with a gain of 0.2\% in mAP, compared with the Baseline.


\section{Conclusion}


We propose a novel plug-and-play detection head optimization approach, namely CEASC, to object detection on drone imagery. It develops the CESC module with CE-GN, which substantially compensates the loss of global context and stabilizes the distribution of foreground. Furthermore, it designs the AMM module to adaptively adjust the mask ratio for distinct foreground areas. Extensive experimental results achieved on VisDrone and UAVDT demonstrate that CEASC remarkably accelerates the inference speed of various base detectors with competitive accuracies.

\section*{Acknowledgment}

This work is partly supported by the National Key R\&D Program of China (2021ZD0110503), the National Natural Science Foundation of China (62022011 and 62202034), the Research Program of State Key Laboratory of Software Development Environment (SKLSDE-2021ZX-04), and the Fundamental Research Funds for the Central Universities.


{\small
\bibliographystyle{ieee_fullname}
\bibliography{egbib}

\begin{thebibliography}{10}\itemsep=-1pt

\bibitem{yolov42020}
Alexey Bochkovskiy, Chien-Yao Wang, and Hong-Yuan~Mark Liao.
\newblock Yolov4: Optimal speed and accuracy of object detection.
\newblock {\em arXiv preprint arXiv:2004.10934}, 2020.

\bibitem{DistillationDetection2017}
Guobin Chen, Wongun Choi, Xiang Yu, Tony Han, and Manmohan Chandraker.
\newblock Learning efficient object detection models with knowledge
  distillation.
\newblock In {\em NeurIPS}, 2017.

\bibitem{mmdetection2019}
Kai Chen, Jiaqi Wang, Jiangmiao Pang, Yuhang Cao, Yu Xiong, Xiaoxiao Li,
  Shuyang Sun, Wansen Feng, Ziwei Liu, Jiarui Xu, et~al.
\newblock Mmdetection: Open mmlab detection toolbox and benchmark.
\newblock {\em arXiv preprint arXiv:1906.07155}, 2019.

\bibitem{UAVDT2018}
Dawei Du, Yuankai Qi, Hongyang Yu, Yifan Yang, Kaiwen Duan, Guorong Li, Weigang
  Zhang, Qingming Huang, and Qi Tian.
\newblock The unmanned aerial vehicle benchmark: Object detection and tracking.
\newblock In {\em ECCV}, 2018.

\bibitem{centernet2019}
Kaiwen Duan, Song Bai, Lingxi Xie, Honggang Qi, Qingming Huang, and Qi Tian.
\newblock Centernet: Keypoint triplets for object detection.
\newblock In {\em ICCV}, 2019.

\bibitem{SACT2017}
Michael Figurnov, Maxwell~D Collins, Yukun Zhu, Li Zhang, Jonathan Huang,
  Dmitry Vetrov, and Ruslan Salakhutdinov.
\newblock Spatially adaptive computation time for residual networks.
\newblock In {\em CVPR}, 2017.

\bibitem{yolox2021}
Zheng Ge, Songtao Liu, Feng Wang, Zeming Li, and Jian Sun.
\newblock Yolox: Exceeding yolo series in 2021.
\newblock {\em arXiv preprint arXiv:2107.08430}, 2021.

\bibitem{RCNN2014}
Ross Girshick, Jeff Donahue, Trevor Darrell, and Jitendra Malik.
\newblock Rich feature hierarchies for accurate object detection and semantic
  segmentation.
\newblock In {\em CVPR}, 2014.

\bibitem{ghostnet2020}
Kai Han, Yunhe Wang, Qi Tian, Jianyuan Guo, Chunjing Xu, and Chang Xu.
\newblock Ghostnet: More features from cheap operations.
\newblock In {\em CVPR}, 2020.

\bibitem{MaskRCNN2017}
Kaiming He, Georgia Gkioxari, Piotr Doll{\'a}r, and Ross Girshick.
\newblock Mask r-cnn.
\newblock In {\em ICCV}, 2017.

\bibitem{ResNet2016}
Kaiming He, Xiangyu Zhang, Shaoqing Ren, and Jian Sun.
\newblock Deep residual learning for image recognition.
\newblock In {\em CVPR}, 2016.

\bibitem{MobileNetV32019}
Andrew Howard, Mark Sandler, Grace Chu, Liang-Chieh Chen, Bo Chen, Mingxing
  Tan, Weijun Wang, Yukun Zhu, Ruoming Pang, Vijay Vasudevan, et~al.
\newblock Searching for mobilenetv3.
\newblock In {\em ICCV}, 2019.

\bibitem{MobileNetV12017}
Andrew~G Howard, Menglong Zhu, Bo Chen, Dmitry Kalenichenko, Weijun Wang,
  Tobias Weyand, Marco Andreetto, and Hartwig Adam.
\newblock Mobilenets: Efficient convolutional neural networks for mobile vision
  applications.
\newblock {\em arXiv preprint arXiv:1704.04861}, 2017.

\bibitem{ufpmpdet2022}
Yecheng Huang, Jiaxin Chen, and Di Huang.
\newblock Ufpmp-det: Toward accurate and efficient object detection on drone
  imagery.
\newblock In {\em AAAI}, 2022.

\bibitem{ccnet2019}
Zilong Huang, Xinggang Wang, Lichao Huang, Chang Huang, Yunchao Wei, and Wenyu
  Liu.
\newblock Ccnet: Criss-cross attention for semantic segmentation.
\newblock In {\em ICCV}, 2019.

\bibitem{BatchNorm2015}
Sergey Ioffe and Christian Szegedy.
\newblock Batch normalization: Accelerating deep network training by reducing
  internal covariate shift.
\newblock In {\em ICML}, 2015.

\bibitem{FocusAndDetect2022}
Onur~Can Koyun, Reyhan~Kevser Keser, {\.I}brahim~Batuhan Akkaya, and
  Beh{\c{c}}et~U{\u{g}}ur T{\"o}reyin.
\newblock Focus-and-detect: A small object detection framework for aerial
  images.
\newblock {\em SPIC}, 104:116675, 2022.

\bibitem{yolov62022}
Chuyi Li, Lulu Li, Hongliang Jiang, Kaiheng Weng, Yifei Geng, Liang Li, Zaidan
  Ke, Qingyuan Li, Meng Cheng, Weiqiang Nie, et~al.
\newblock Yolov6: A single-stage object detection framework for industrial
  applications.
\newblock {\em arXiv preprint arXiv:2209.02976}, 2022.

\bibitem{DMNet2020}
Changlin Li, Taojiannan Yang, Sijie Zhu, Chen Chen, and Shanyue Guan.
\newblock Density map guided object detection in aerial images.
\newblock In {\em CVPR Workshops}, 2020.

\bibitem{GFLv22021}
Xiang Li, Wenhai Wang, Xiaolin Hu, Jun Li, Jinhui Tang, and Jian Yang.
\newblock Generalized focal loss v2: Learning reliable localization quality
  estimation for dense object detection.
\newblock In {\em CVPR}, 2021.

\bibitem{GFLv12020}
Xiang Li, Wenhai Wang, Lijun Wu, Shuo Chen, Xiaolin Hu, Jun Li, Jinhui Tang,
  and Jian Yang.
\newblock Generalized focal loss: Learning qualified and distributed bounding
  boxes for dense object detection.
\newblock In {\em NeurIPS}, 2020.

\bibitem{RetinaNet2017}
Tsung-Yi Lin, Priya Goyal, Ross Girshick, Kaiming He, and Piotr Doll{\'a}r.
\newblock Focal loss for dense object detection.
\newblock In {\em ICCV}, 2017.

\bibitem{COCO2014}
Tsung-Yi Lin, Michael Maire, Serge Belongie, James Hays, Pietro Perona, Deva
  Ramanan, Piotr Doll{\'a}r, and C~Lawrence Zitnick.
\newblock Microsoft coco: Common objects in context.
\newblock In {\em ECCV}, 2014.

\bibitem{groupfisherprune2021}
Liyang Liu, Shilong Zhang, Zhanghui Kuang, Aojun Zhou, Jing-Hao Xue, Xinjiang
  Wang, Yimin Chen, Wenming Yang, Qingmin Liao, and Wayne Zhang.
\newblock Group fisher pruning for practical network compression.
\newblock In {\em ICML}, 2021.

\bibitem{networkslimming2017}
Zhuang Liu, Jianguo Li, Zhiqiang Shen, Gao Huang, Shoumeng Yan, and Changshui
  Zhang.
\newblock Learning efficient convolutional networks through network slimming.
\newblock In {\em ICCV}, 2017.

\bibitem{ShufflenetV22018}
Ningning Ma, Xiangyu Zhang, Hai-Tao Zheng, and Jian Sun.
\newblock Shufflenet v2: Practical guidelines for efficient cnn architecture
  design.
\newblock In {\em ECCV}, 2018.

\bibitem{pytorch2019}
Adam Paszke, Sam Gross, Francisco Massa, Adam Lerer, James Bradbury, Gregory
  Chanan, Trevor Killeen, Zeming Lin, Natalia Gimelshein, Luca Antiga, et~al.
\newblock Pytorch: An imperative style, high-performance deep learning library.
\newblock In {\em NeurIPS}, 2019.

\bibitem{thundernet2019}
Zheng Qin, Zeming Li, Zhaoning Zhang, Yiping Bao, Gang Yu, Yuxing Peng, and
  Jian Sun.
\newblock Thundernet: Towards real-time generic object detection on mobile
  devices.
\newblock In {\em ICCV}, 2019.

\bibitem{Faster-RCNN2015}
Shaoqing Ren, Kaiming He, Ross Girshick, and Jian Sun.
\newblock Faster r-cnn: Towards real-time object detection with region proposal
  networks.
\newblock In {\em NeurIPS}, 2015.

\bibitem{MobileNetV22018}
Mark Sandler, Andrew Howard, Menglong Zhu, Andrey Zhmoginov, and Liang-Chieh
  Chen.
\newblock Mobilenetv2: Inverted residuals and linear bottlenecks.
\newblock In {\em CVPR}, 2018.

\bibitem{DynamicHead2020}
Lin Song, Yanwei Li, Zhengkai Jiang, Zeming Li, Hongbin Sun, Jian Sun, and
  Nanning Zheng.
\newblock Fine-grained dynamic head for object detection.
\newblock In {\em NeurIPS}, 2020.

\bibitem{efficientnet2019}
Mingxing Tan and Quoc Le.
\newblock Efficientnet: Rethinking model scaling for convolutional neural
  networks.
\newblock In {\em ICML}, 2019.

\bibitem{FCOS2019}
Zhi Tian, Chunhua Shen, Hao Chen, and Tong He.
\newblock Fcos: Fully convolutional one-stage object detection.
\newblock In {\em ICCV}, 2019.

\bibitem{InstanceNorm2016}
Dmitry Ulyanov, Andrea Vedaldi, and Victor Lempitsky.
\newblock Instance normalization: The missing ingredient for fast stylization.
\newblock {\em arXiv preprint arXiv:1607.08022}, 2016.

\bibitem{DynamicConvolution2020}
Thomas Verelst and Tinne Tuytelaars.
\newblock Dynamic convolutions: Exploiting spatial sparsity for faster
  inference.
\newblock In {\em CVPR}, 2020.

\bibitem{yolov72022}
Chien-Yao Wang, Alexey Bochkovskiy, and Hong-Yuan~Mark Liao.
\newblock Yolov7: Trainable bag-of-freebies sets new state-of-the-art for
  real-time object detectors.
\newblock {\em arXiv preprint arXiv:2207.02696}, 2022.

\bibitem{nasfcos2020}
Ning Wang, Yang Gao, Hao Chen, Peng Wang, Zhi Tian, Chunhua Shen, and Yanning
  Zhang.
\newblock Nas-fcos: Fast neural architecture search for object detection.
\newblock In {\em CVPR}, 2020.

\bibitem{cbam2018}
Sanghyun Woo, Jongchan Park, Joon-Young Lee, and In~So Kweon.
\newblock Cbam: Convolutional block attention module.
\newblock In {\em ECCV}, 2018.

\bibitem{GN2018}
Yuxin Wu and Kaiming He.
\newblock Group normalization.
\newblock In {\em ECCV}, 2018.

\bibitem{SpatiallyAdaptiveInference2020}
Zhenda Xie, Zheng Zhang, Xizhou Zhu, Gao Huang, and Stephen Lin.
\newblock Spatially adaptive inference with stochastic feature sampling and
  interpolation.
\newblock In {\em ECCV}, 2020.

\bibitem{SECOND2018}
Yan Yan, Yuxing Mao, and Bo Li.
\newblock Second: Sparsely embedded convolutional detection.
\newblock {\em Sensors}, 18(10):3337, 2018.

\bibitem{QueryDet2022}
Chenhongyi Yang, Zehao Huang, and Naiyan Wang.
\newblock Querydet: Cascaded sparse query for accelerating high-resolution
  small object detection.
\newblock In {\em CVPR}, 2022.

\bibitem{ClusDet2019}
Fan Yang, Heng Fan, Peng Chu, Erik Blasch, and Haibin Ling.
\newblock Clustered object detection in aerial images.
\newblock In {\em ICCV}, 2019.

\bibitem{FGD2022}
Zhendong Yang, Zhe Li, Xiaohu Jiang, Yuan Gong, Zehuan Yuan, Danpei Zhao, and
  Chun Yuan.
\newblock Focal and global knowledge distillation for detectors.
\newblock In {\em CVPR}, 2022.

\bibitem{slimyolov32019}
Pengyi Zhang, Yunxin Zhong, and Xiaoqiong Li.
\newblock Slimyolov3: Narrower, faster and better for real-time uav
  applications.
\newblock In {\em ICCV Workshops}, 2019.

\bibitem{ATSS2020}
Shifeng Zhang, Cheng Chi, Yongqiang Yao, Zhen Lei, and Stan~Z Li.
\newblock Bridging the gap between anchor-based and anchor-free detection via
  adaptive training sample selection.
\newblock In {\em CVPR}, 2020.

\bibitem{ShufflenetV12018}
Xiangyu Zhang, Xinyu Zhou, Mengxiao Lin, and Jian Sun.
\newblock Shufflenet: An extremely efficient convolutional neural network for
  mobile devices.
\newblock In {\em CVPR}, 2018.

\bibitem{FSAF2019}
Chenchen Zhu, Yihui He, and Marios Savvides.
\newblock Feature selective anchor-free module for single-shot object
  detection.
\newblock In {\em CVPR}, 2019.

\bibitem{VisDrone2018}
Pengfei Zhu, Longyin Wen, Xiao Bian, Haibin Ling, and Qinghua Hu.
\newblock Vision meets drones: A challenge.
\newblock {\em arXiv preprint arXiv:1804.07437}, 2018.

\end{thebibliography}
}

\clearpage
\appendix

\renewcommand{\thefigure}{\Alph{figure}}
\renewcommand{\thesection}{\Alph{section}}
\renewcommand{\thetable}{\Alph{table}}
\setcounter{table}{0}
\setcounter{figure}{0}
\setcounter{section}{0}

\section*{Supplementary Material}

In this document, we provide more implementation details, additional ablation studies as well as more visualization results. 

\section{More Implementation Details}\label{sec:impl}
In Table~\ref{tab:sota_diff_basedet} of the main body, we evaluate the performance of our approach with various base detectors, including RetinaNet, FSAF, Faster-RCNN besides GFL V1. Since both RetinaNet and FSAF are one-stage detectors as GFL V1, we adopt the same setting as used by GFL V1. Regarding the two-stage Faster-RCNN detector, we follow \cite{QueryDet2022}, modifying its RPN head to 4 Convolution-GN-ReLU layers instead of 1 convolution layer and using 256 feature channels, which proves effective in balancing the accuracy and efficiency \cite{QueryDet2022}. 

In Table~\ref{tab:sota_on_visdrone} of the main body, we compare our network to MobileNet V2~\cite{MobileNetV22018}, ShuffleNet V2~\cite{ShufflenetV22018} and QueryDet~\cite{QueryDet2022}. We select the feature maps generated from the layers (2, 4, 6) and (0, 1, 2) as the input of FPN for MobileNet V2 and ShuffleNet V2, respectively. Regarding QueryDet, we re-implement it by using the same setting for fair comparison. Particularly, we utilize the unified input size as 1,333$\times$800 and omit the calculation of the $P_{2}$ layer. 

\section{Additional Ablation Studies}\label{sec:ab}
We show additional results by our approach using different residual structures, accelerating strategies, context clues and training epochs.

\subsection{On Residual Structures}
As shown in Fig.~\ref{fig:framework} of the main body, we adopt a residual structure to compensate the loss of contextual information due to sparse convolutions. We therefore compare the performance of the proposed CEASC network by using different residual structures, including: 1) ``w.o. Res.'' without using the residual structure; 2) ``Focal Res.'' using the raw input for skip connection, \emph{i.e.} $\mathbf{F}:=\mathbf{F}+\mathbf{X}$; and 3) ``Ours'' using the global contextual feature for skip connection, \emph{i.e.} $\mathbf{F}:=\mathbf{F}+\mathbf{G}$.

As displayed in Table \ref{tab:app_ab_res}, our approach reaches the best performance, highlighting its advantage in capturing global context. 

\subsection{On Acceleration Strategies}
\begin{figure*}[!thb]
\centering
\includegraphics[width=\linewidth]{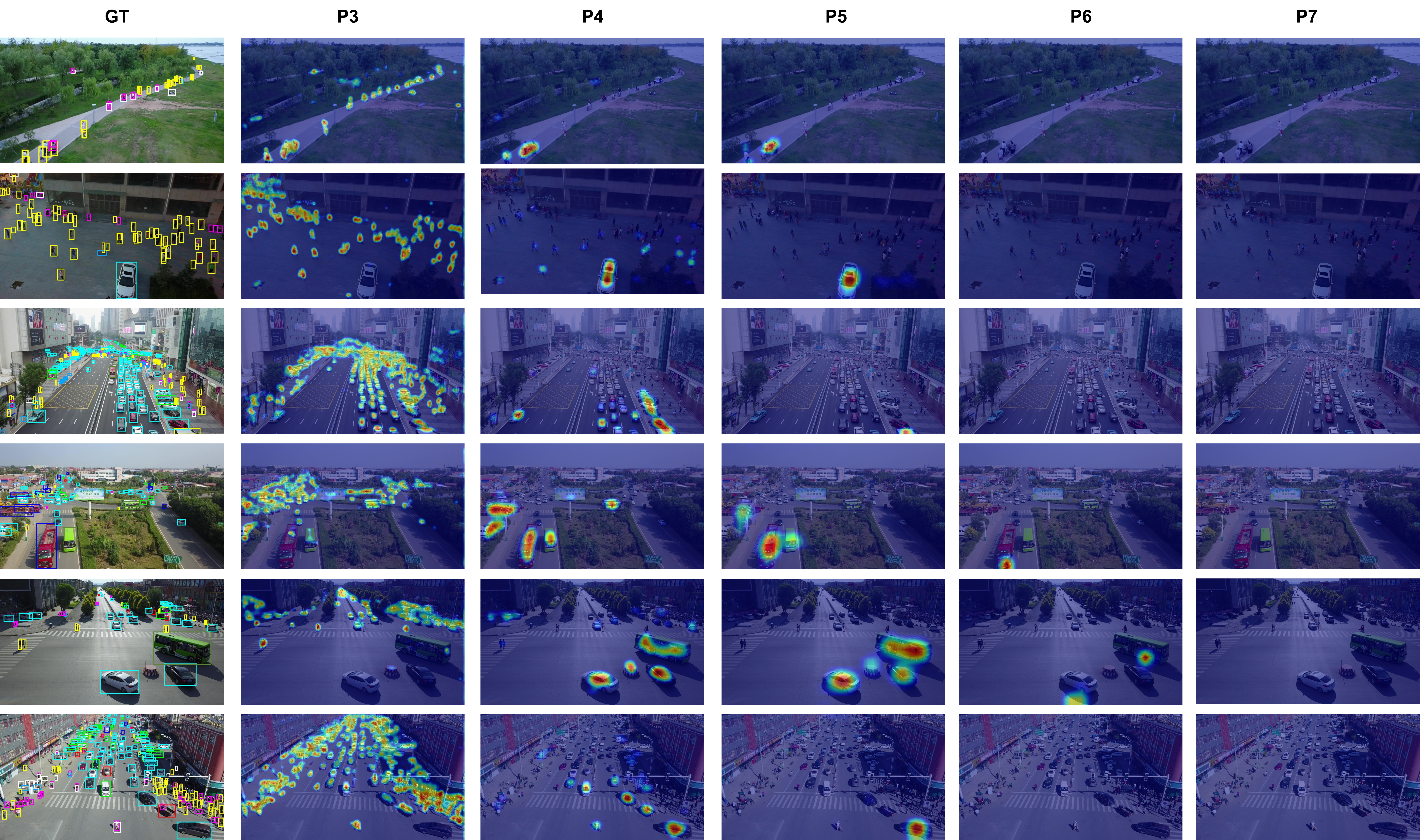}
\caption{Visualization of dynamic masks estimated by AMM at different layers (from `P3' to `P7') in FPN of GFL V1 on VisDrone. Highlighted areas are activated for computation.}
\label{fig:app_visdrone}
\end{figure*}

In Table~\ref{tab:sota_on_visdrone} of the main body, we compare our approach with the state-of-the-art lightweight models for drone-based object detection. Here, we carry out the ablation study on our approach using distinct acceleration strategies including: 1) ``FPN on P3+P4 only'' that only adopts the P3 and P4 layers for FPN, since the P5-P7 layers are unlikely to be activated for sparse convolutions as observed in Fig~\ref{fig:vis} and Fig.~\ref{fig:app_visdrone};  and 2) ``DWS'' that utilizes the Depth-Wise Separable (DWS) convolution as in MobileNet, instead of the normal 3 $\times$ 3 convolution in the ``Baseline'', \emph{i.e.} the original GFL V1 detector.

The results are summarized in Table \ref{tab:app_ab_acc}, indicating that our approach outperforms the counterparts both in accuracy and efficiency, due to the sparse convolutions optimized by context-enhancement and adaptive multi-layer masking.

\begin{table}[!t]
    \centering
    \begin{tabular}{c|ccc|cc}
    \hline
    Method & mAP & $\text{AP}_{50}$ & $\text{AP}_{75}$ & GFLOPs & FPS \\ \hline
    w.o. Res. & 28.4 & 50.5 & 28.1 & 150.58 & 21.32 \\
    Focal Res. & 28.1 & 49.8 & 27.4 & 153.52 & 19.98 \\
    \textbf{Ours} & \textbf{28.7} & \textbf{50.7} & \textbf{28.4} & \textbf{150.18} & \textbf{21.55} \\  \hline
    \end{tabular}
    \caption{Comparison in terms of mAP (\%) and GFLOPs/FPS with different residual structures on VisDrone.}
    \label{tab:app_ab_res}
\end{table}

\begin{table}[!t]
    \centering
    \resizebox{1\linewidth}{!}{
    \begin{tabular}{c|ccc|cc}
    \hline
    Method & mAP & $\text{AP}_{50}$ & $\text{AP}_{75}$ & GFLOPs & FPS \\ \hline
    Baseline & 28.4 & 50.0 & 27.8 & 524.95 & 13.46 \\
    FPN on P3+P4 only & 28.1 & 49.9 & 27.5 & 494.42 & 15.63 \\
    DWS & 24.5 & 43.2 & 23.9 & 157.74 & 20.84 \\
    \textbf{Ours} & \textbf{28.7} & \textbf{50.7} & \textbf{28.4} & \textbf{150.18} & \textbf{21.55} \\  \hline
    \end{tabular}}
    \caption{Comparison in terms of mAP (\%) and GFLOPs/FPS with different acceleration strategies on VisDrone.}
    \label{tab:app_ab_acc}
\end{table}

\begin{table}[!t]
    \centering
    \resizebox{1\linewidth}{!}{
    \begin{tabular}{c|ccc|cc}
    \hline
    Method & mAP & $\text{AP}_{50}$ & $\text{AP}_{75}$ & GFLOPs & FPS \\ \hline
    Baseline & 28.4 & 50.0 & 27.8 & 524.95 & 13.46 \\
    Interpolation\cite{SpatiallyAdaptiveInference2020} & 26.9 & 48.5 & 26.1 & 212.15 & 13.80 \\
    \textbf{Ours} & \textbf{28.7} & \textbf{50.7} & \textbf{28.4} & \textbf{150.18} & \textbf{21.55} \\  \hline
    \end{tabular}}
    \caption{Comparison in terms of mAP (\%) and GFLOPs/FPS with different context clues on VisDrone.}
    \label{tab:app_ab_context}
\end{table}

\begin{figure*}[!thb]
\centering
\includegraphics[width=\linewidth]{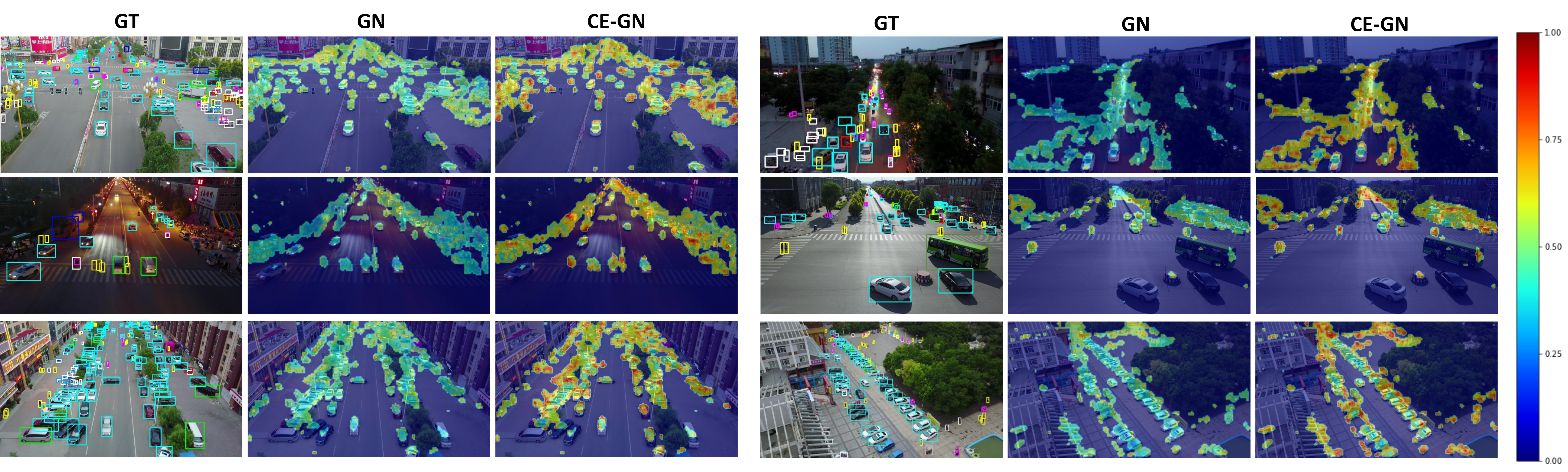}
\caption{Visualization of correlation between features generated by dense convolutions and sparse convolutions with distinct normalization schemes on VisDrone.}
\label{fig:app_ce-gn}
\end{figure*}

\begin{table*}[!t]
    \centering
    \normalsize
    \begin{tabular}{c|c|ccc|cccc|cc}
    \hline
    Epoch & Method & mAP & $\text{AP}_{50}$ & $\text{AP}_{75}$ & $\text{AR}_{1}$ & $\text{AR}_{10}$ & $\text{AR}_{100}$ & $\text{AR}_{500}$ & GFLOPs & FPS \\ \hline
    \multirow{2}*{12}        & Baseline & 27.8 & 49.2 & 27.3 & 0.63 & 6.27 & 34.7 & 44.2 & 524.95 & 13.48 \\
                                 & \textbf{Ours (CEASC)} & 27.8 & \textbf{49.3} & \textbf{27.4} & \textbf{0.67} & \textbf{6.47} & \textbf{34.8} & \textbf{44.4} & \textbf{151.93} & \textbf{21.52} \\\hline
    \multirow{2}*{15}   & Baseline & 28.4 & 50.0 & 27.8 & 0.62 & 6.36 & 35.6 & 44.9 & 524.95 & 13.46 \\\
                                 & \textbf{Ours (CEASC)} & \textbf{28.7} & \textbf{50.7} & \textbf{28.4} & \textbf{0.65} & \textbf{6.56} & 35.6 & \textbf{45.0} & \textbf{150.18} & \textbf{21.55}\\\hline
    \multirow{2}*{24}   & Baseline  & 28.9 & 50.9 & 28.4 & \textbf{0.72} & 6.53 & 35.7 & 45.2 & 524.95 & 13.41 \\
                                 & \textbf{Ours (CEASC)}  & \textbf{29.1} & \textbf{51.3} & \textbf{28.7} & 0.70 & \textbf{6.90} & \textbf{36.0} & \textbf{45.4} & \textbf{151.42} & \textbf{21.49} \\\hline
    \end{tabular}
    \caption{Comparison in terms of AP/AR (\%) and GFLOPs/FPS with the GFL V1 base detector using different training epochs on VisDrone.}
    \label{tab:app_ab_epoch}
\end{table*}

\begin{figure*}[!thb]
\centering
\includegraphics[width=\linewidth]{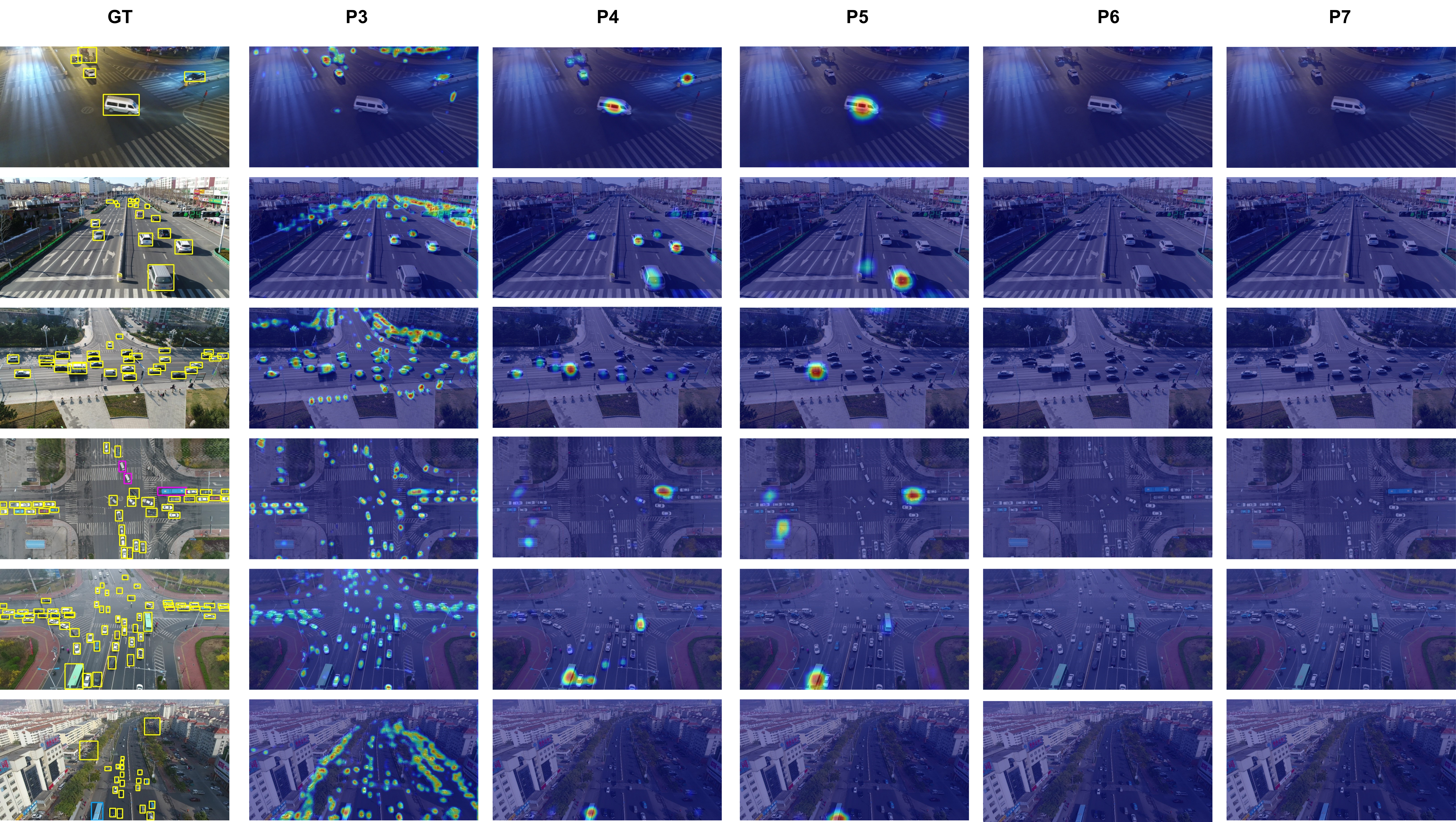}
\caption{Visualization of dynamic masks estimated by AMM at different layers (from `P3' to `P7') in FPN of GFL V1 on UAVDT. Highlighted areas are activated for computation.}
\label{fig:app_uavdt}
\end{figure*}


\subsection{On Context Clues}
In Sec.~\ref{sec:CE} of the main body, we mention an interpolation method to generate ignored pixels from focal areas~\cite{SpatiallyAdaptiveInference2020}, and an ablation study is conducted for comparison. The results in Table~\ref{tab:app_ab_context} reveal that interpolation incurs a drop on the accuracy but consumes more computations.

\subsection{On Training Epochs}

In the literature, some studies train their models for varying epochs (\emph{e.g.} 12 or 24). We thus provide more results by using such numbers of training epochs in addition to 15 adopted in this work. As displayed in Table \ref{tab:app_ab_epoch}, our approach consistently boosts the performance by a large margin with different training epochs. When more training epochs are used, our approach reaches a higher accuracy, where 15 is a good trade-off. 

\subsection{More Visualized Results}
More results are visualized in Fig.~\ref{fig:app_ce-gn} as supplements to Fig.~\ref{fig:ce-gn} of the main body. The features normalized by CE-GN have higher correlation with dense convolutions than GN, indicating that CE-GN enhances focal features with the assistance of global context.

In Fig.~\ref{fig:app_visdrone} and Fig.~\ref{fig:app_uavdt}, we visualize more results as supplements to Fig.~\ref{fig:vis} of the main body. As illustrated, the mask generated by our approach well covers foreground areas, indicating that sparse convolutions spend most computations on foreground, thus promoting the efficiency without sacrificing much precision.

\end{document}